\documentclass[10pt, conference]{IEEEtran}
\usepackage{amsmath,amssymb,amsfonts}
\usepackage{graphicx}
\usepackage{pifont}

\usepackage{amsthm}
\usepackage{xcolor}
\usepackage[figuresright]{rotating}

\usepackage{graphicx}
\graphicspath{{/}{fig/}}

\usepackage{array}
\usepackage{textcomp}
\usepackage{xcolor}
\usepackage{multirow}
\usepackage{booktabs}

\usepackage{mathtools}
\usepackage{breqn}
\usepackage{float}

\usepackage{pgfplots}
\pgfplotsset{compat=1.7}
\usepgfplotslibrary{groupplots}

\usepackage{caption}
\usepackage{subcaption}

\usepackage{multirow,tabularx}
\usepackage{hyperref}
\usepackage[ancient]{flushend}
\usepackage{algorithmic}
\usepackage[vlined, ruled, shortend]{algorithm2e}
\newlength\figureheight
\newlength\figurewidth
\SetAlCapNameFnt{\footnotesize}
\SetAlCapFnt{\footnotesize}

\title{
    Analyzing General-Purpose Deep-Learning Detection and Segmentation Models with Images from a Lidar as a Camera Sensor \\
}

\author{
    \IEEEauthorblockN{
        \vspace{1em}
        Yu Xianjia\IEEEauthorrefmark{2},
        Sahar Salimpour\IEEEauthorrefmark{2},
        Jorge Peña Queralta\IEEEauthorrefmark{2},
        Tomi Westerlund\IEEEauthorrefmark{2}
    }
    \IEEEauthorblockA{
        \normalsize
        \IEEEauthorrefmark{2}\href{https://tiers.utu.fi}{Turku Intelligent Embedded and Robotic Systems (TIERS) Lab, University of Turku, Finland}.\\
        Emails: \textsuperscript{1}\{xianjia.yu, sahars, jopequ,  tovewe\}@utu.fi\\[+6pt]
    }
}

\begin{document}

\maketitle
\thispagestyle{empty}
\pagestyle{empty}

%%%%%%%%%%%%%%%%%%%%%%%%%%%%%%%%%%%%%%%%%%%%%%
%%                                          %%
%%           ABSTRACT AND TITLE             %%
%%                                          %%
%%%%%%%%%%%%%%%%%%%%%%%%%%%%%%%%%%%%%%%%%%%%%%
\begin{abstract}

Over the last decade, robotic perception algorithms have significantly benefited from the rapid advances in deep learning (DL). Indeed, a significant amount of the autonomy stack of different commercial and research platforms relies on DL for situational awareness, especially vision sensors. This work explores the potential of general-purpose DL perception algorithms, specifically detection and segmentation neural networks, for processing image-like outputs of advanced lidar sensors. Rather than processing the three-dimensional point cloud data, this is, to the best of our knowledge, the first work to focus on low-resolution images with 360\textdegree field of view obtained with lidar sensors by encoding either depth, reflectivity, or near-infrared light in the image pixels. We show that with adequate preprocessing, general-purpose DL models can process these images, opening the door to their usage in environmental conditions where vision sensors present inherent limitations. We provide both a qualitative and quantitative analysis of the performance of a variety of neural network architectures. We believe that using DL models built for visual cameras offers significant advantages due to the much wider availability and maturity compared to point cloud-based perception.


\end{abstract}

\IEEEpeerreviewmaketitle

%%%%%%%%%%%%%%%%%%%%%%%%%%%%%%%%%%%%%%%%%%%%%%
%%                                          %%
%%                SECTIONS                  %%
%%                                          %%
%%%%%%%%%%%%%%%%%%%%%%%%%%%%%%%%%%%%%%%%%%%%%%

\section{Introduction}
\label{sec:intro}

Autonomous mobile robots and self-driving cars use a variety of sensors for ensuring a high level of situational awareness~\cite{fan2019key}. For instance, the Autoware project, representing the state-of-the-art autonomous cars, relies on 3D lidars for key perception components~\cite{kato2018autoware}. Multiple aerial robotic solutions also utilize lidars for autonomous flight in complex environments~\cite{liu2022large}. Some of the critical characteristics of lidars that motivate their adoption across application fields include their long-range and the accuracy of the geometric data they output.

    % In recent years, there has been rapid development of numerous new sensor technologies, including lidar sensors.  
Lidar point cloud data features 360\textdegree three-dimensional high spatial resolution data but often a limited vertical field of view. Advanced sensors have vertical resolutions that typically range from 30\textdegree to 90\textdegree~\cite{maksymova2018review}.  %and wide vertical field of view by analyzing laser's energy density returns from reflecting surfaces. 
As lidars measure the time of flight of a laser signal to objects in the environment, they are
%Therefore, it is 
not influenced by changes in light such as darkness and daylight. In several studies, lidar point cloud data and image data have been used together in a variety of computer vision tasks, such as 3D object detection\cite{yoo20203d, qingqing2021adaptive, zhong2021survey, xianjia2021cooperative}. However, while lidar odometry, localization and mapping are at the pinnacle of autonomous technology~\cite{li2020multi}, the processing of point cloud data for object detection or semantic scene segmentation is not as mature as the algorithms, and machine learning (ML) approaches for vision sensors~\cite{cui2021deep, qingqing2019monocular}.

\begin{figure}[t]
    \centering
    
    \begin{subfigure}{.48\textwidth}
        \centering
        \includegraphics[width=\textwidth,height=.29\textwidth]{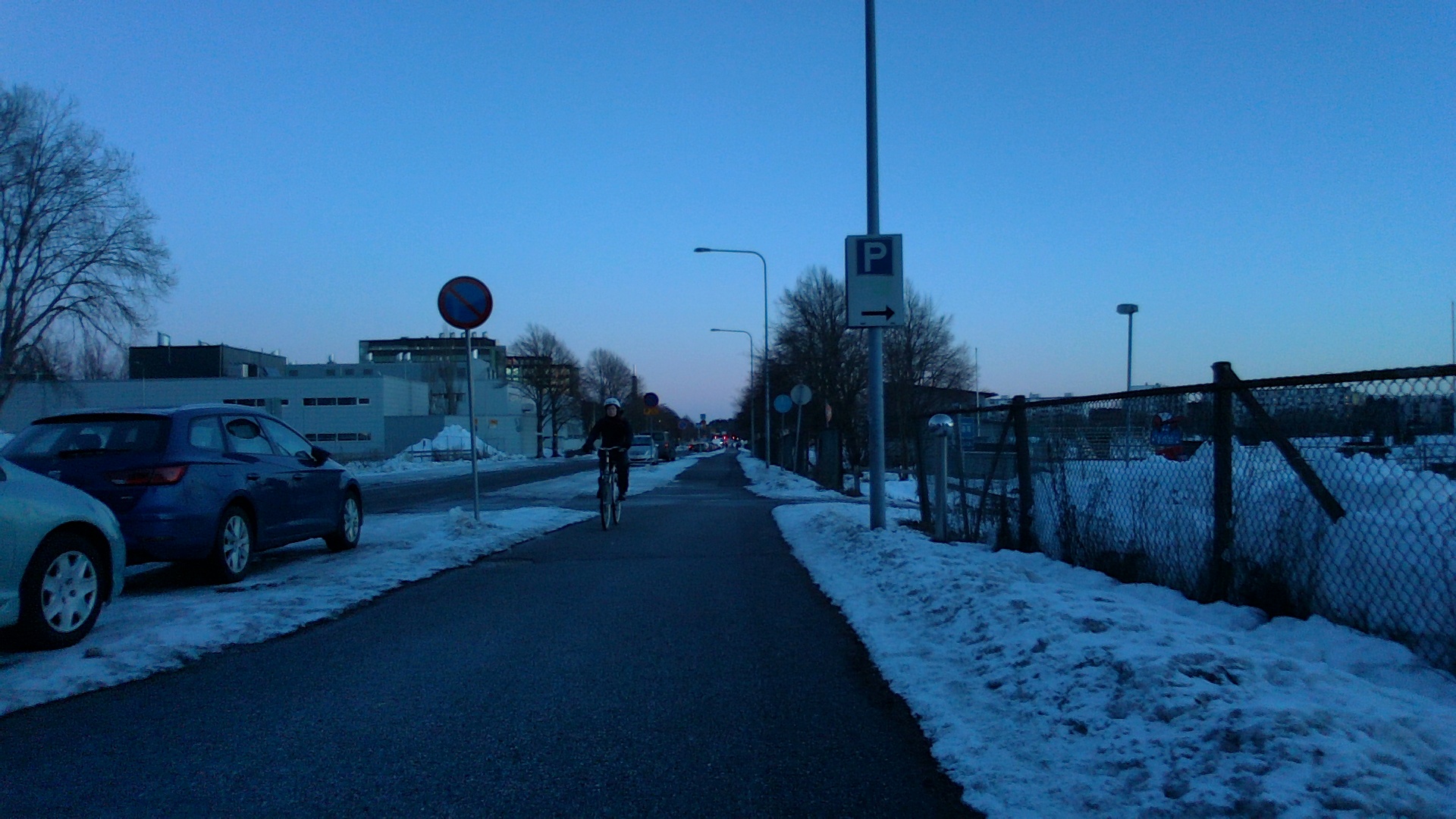}
        \caption{Outdoors (RGB)}
        \label{fig:sample_outdoors_rgb}
    \end{subfigure}
    
    \vspace{1em}
    \begin{subfigure}{.48\textwidth}
        \centering
        \includegraphics[width=\textwidth, height=0.29\textwidth]{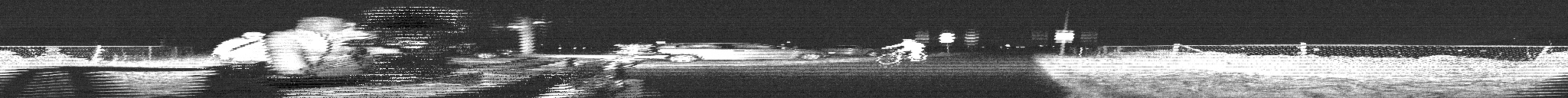}
        \caption{Outdoors (Lidar)}
        \label{fig:sample_indoors_lidar}
    \end{subfigure}
    
    \vspace{1em}
    \begin{subfigure}{.48\textwidth}
        \centering
        \includegraphics[width=\textwidth, height=0.29\textwidth]{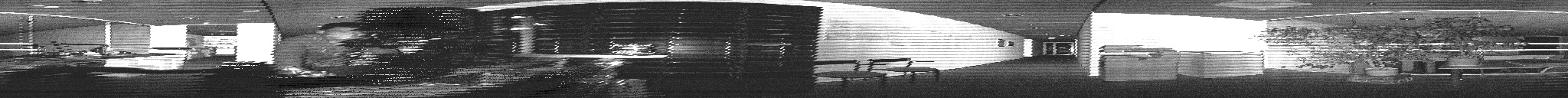}
        \caption{Indoors (Lidar)}
        \label{fig:sample_outdoors_lidar}
    \end{subfigure}
    
    \vspace{1em}
    \begin{subfigure}{.48\textwidth}
        \centering
        \includegraphics[width=\textwidth, height=.3\textwidth]{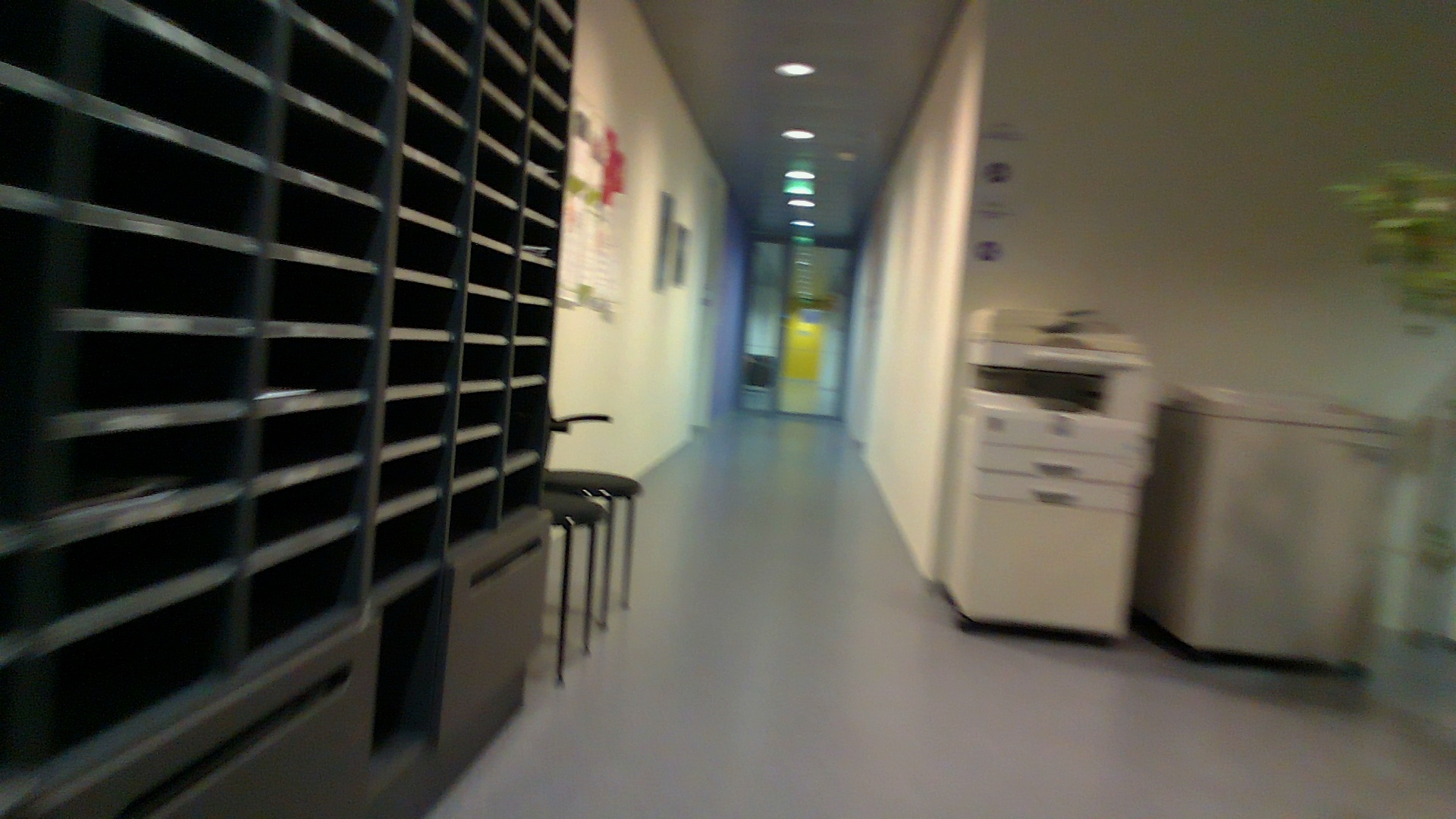}
        \caption{Indoors (RGB)}
        \label{fig:sample_indoors_rgb}
    \end{subfigure}
    \caption{Samples of images utilized in this work. The outdoors sample includes a bicycle that is seen in both the RGB and lidar data, as well as several cars. In both cases, a person behind the sensors does appear in the 360\textdegree ~lidar image but not in the RGB frame.}
    \label{fig:data_samples}
\end{figure}

Deep learning (DL) has revolutionized computer vision over the last decade. Within the robotics field, from advanced perception~\cite{pierson2017deep} to novel end-to-end control architectures based on deep reinforcement learning~\cite{zhao2020sim}, and including odometry and localization~\cite{qingqing2019offloading}. We are in this work particularly interested in DL models for object detection and instance segmentation, both of which are cornerstones to embedding intelligence into autonomous robots and enabling high degrees of situational awareness~\cite{queralta2020collaborative}. Even though most of the work in DL-based perception has focused on images and vision sensors, DL applications to lidar data include voxel-based object detection or point cloud segmentation~\cite{li2020deep}. The literature also includes multiple examples of lidar and camera fusion for producing coloured point clouds or more robust behaviour, e.g., when segmenting roads in self-driving cars~\cite{caltagirone2019lidar}. These works, however, focus on point cloud lidar data~\cite{li2020deep}, while we explore the potential also to leverage them as camera-like sensors. Only recently, such potential has been identified~\cite{tsiourva2020lidar}, but the existing literature lacks a more in-depth analysis of the potential of images captured from lidar sensors. A sample of the data used in this work is shown in Fig.~\ref{fig:data_samples}.

% \red{[One paragraph here about DL-based object perception and scene segmentation with both cameras and lidar. Maybe a couple lines about fusion of both sensors (e.g., colored point clouds].}

Albeit the higher cost of lidar at the moment, compared with passive visual sensors, lidars are inherently more robust to adverse weather conditions and low-visibility environments. They are also a standard part of most of today's self-driving autonomy stacks. Therefore, it comes at no extra cost to leverage their vision-like capabilities in addition to processing the three-dimensional point cloud data.

he main contribution of this work is the analysis of the performance of a variety of DL-based visual perception models in lidar camera data. We assess the viability of applying object detection and instance segmentation models to low-resolution, 360\textdegree images from two different Ouster lidars with different fields of view and range. On the object detection side, we utilize both one-stage detectors (YOLOv5 and YOLOx) and two-stage detectors (Faster R-CNN and Mask R-CNN). For semantic instance segmentation, we study the performance of HRNet, PointRend and Mask R-CNN.

The remainder of this document is structured as follows. In Section 2, we overview the literature in DL perception, lidar-based object detection and segmentation, and fusion of vision and lidar sensors. Section 3 then covers the hardware and software methods utilized through our work. In Section 4, we report experimental results, discussing the potential of this type of sensor data in Section 5. Finally, Section 6 concludes the work and outlines future research directions.

% \newpage
\section{Related work}
\label{sec:related}

The literature in the processing of low-resolution lidar-based images is scarce. In~\cite{angus2018lidar}, Ouster's CEO introduces the technology, showcasing the performance of the car and road segmentation using a re-trained DL model with a video. The author also comments on the potential for using this data as an input to a pre-trained network from DeTone et al.'s SuperPoint project for odometry estimations. However, in both cases, the code is not available, and neither are quantitative results shown. Minimal research has been carried out in this direction to the best of our knowledge. In~\cite{tsiourva2020lidar}, Tsiourva et al. analyzed the potential of the same Ouster lidar sensors that we study for saliency detection. This work already demonstrates the more consistent performance and data quality in adverse environments (e.g., rainy weather). We further analyze DL-based perception performance beyond essential computer vision preprocessing such as saliency detection. 

Through the rest of this section, we review the current research directions and the state-of-the-art in lidar-based perception and fusion with cameras, DL-based object detection and segmentation, and fusion of lidar and camera data.

\subsection{Lidar-based perception}

Lidar data provides accurate and reliable depth and geometric information and is a crucial component of various kinds of perception tasks, such as 3D mapping, localization, and object detection~\cite{li2020multi}.

% Lidar data, which provides reliable 1D, 2D, and 3D depth geometry information, has become a significant component of various kinds of perception tasks, such as 3D mapping, localization, and object detection.

There have been many studies carried out on the detection and localization tasks using lidar point cloud data~\cite{zhou2018voxelnet,li20173d}. In most cases, however, current techniques are based on a fusion of both camera and lidar data~\cite{pang2020clocs, wen2021fast, li2021openstreetmap}. In~\cite{caltagirone2019lidar} different fusion methods were applied to detect roads with lidar and camera data. In addition, several studies have been utilizing lidar and camera data to detect pedestrians and vehicles the self-driving systems~\cite{schlosser2016fusing, asvadi2018multimodal}.

\subsection{Deep Learning Based Object Detection}

Object detection has been among the most trivial tasks in computer vision applications. This task has been extensively explored in a wide range of technological advances in recent years, including autonomous driving, identity detection, medical applications, and robotics. In most state-of-the-art object detection methods, deep learning neural network models are used as the backbone to extract features, classify objects and identify their locations~\cite{pierson2017deep, queralta2020collaborative}.

The most popular types of detectors are YOLO~\cite{redmon2016you} (You only look once) and various versions of it~\cite{redmon2017yolo9000,redmon2018yolov3}, RetinaNet~\cite{lin2017focal}, SSD~\cite{liu2016ssd} (Single Shot MultiBox Detector, R-CNN~\cite{girshick2014rich} (Region-CNN) and its extensions, and Mask R-CNN\cite{he2017mask}.

A representative example appears in~\cite{li20173d}, where authors proposed a 3D fully convolutional network based on DenseBox for 3D detection and localization of vehicles from lidar point cloud data. As described in~\cite{kim2019advanced}, RGB camera data and LiDAR point cloud data were combined to enhance the object detection performance in real-time by using a weighted-mean YOLO algorithm. In other approaches, point cloud data was converted into bird's eye view images and then fused with front-facing camera images using multi-view 3D networks to predict 3D bounding boxes~\cite{8100174, ku2018joint}.

\subsection{Deep Learning Based Instance and Semantic Segmentation}

In\cite{geng2020deep} a dual-modal instance segmentation deep neural network based on the architectures of the RetinaNet and Mask R-CNN networks was developed for object segmentation using the RGB and Lidar pixel-level images. The authors in~\cite{wu2018squeezeseg} transformed the 3D Lidar point clouds to 2D grids representations by applying a spherical projection. Then, the SqueezeSeg model derived from SqueezeNet was developed for semantic segmentation of the obtained range images. Alternatively, in~\cite{imad2021transfer} the authors propose a transfer learning model based on MobileNetv2 for semantic segmentation of birds-eye-view representation of the 3D point cloud data.

\begin{figure*}[t]
    \centering
    \includegraphics[width=.65\textwidth]{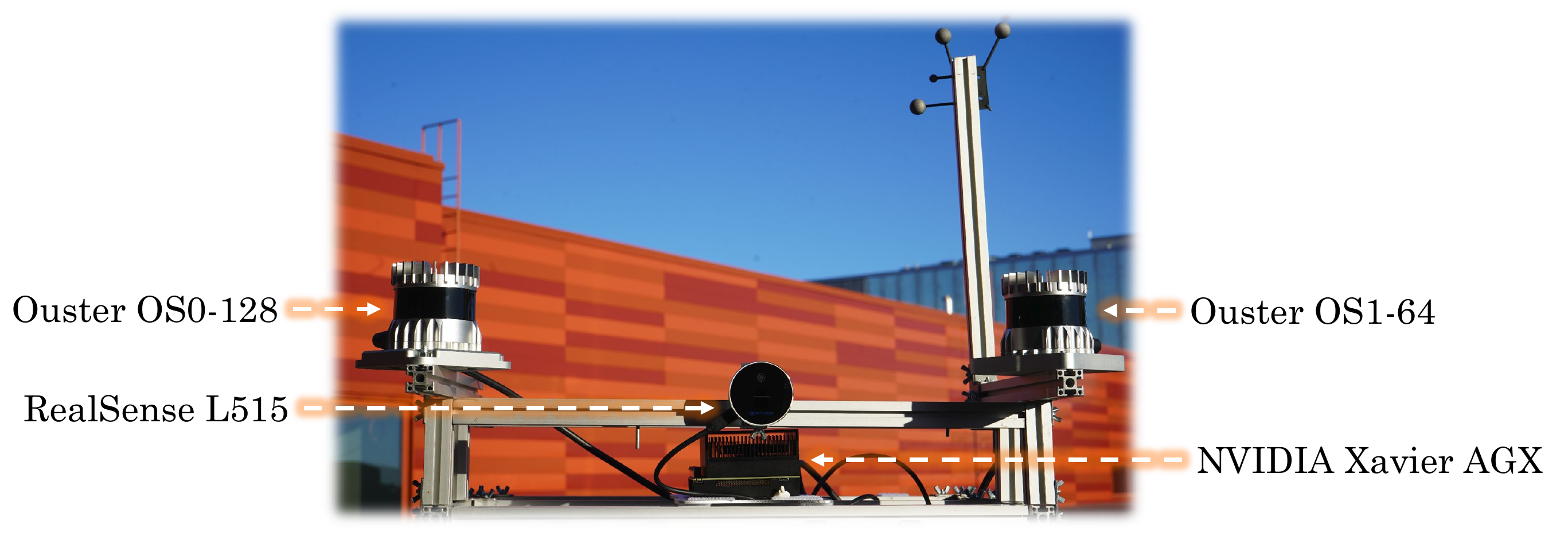}
    \caption{Equipment utilized for data acquisition.}
    \label{fig:hardware_frame}
\end{figure*}

% \newpage
\section{Methodology}

This section covers the hardware and methods utilized in our study. We describe the sensors utilized for data acquisition as well as the different DL model architectures.

\subsection{Hardware}

The equipment for data acquisition consists of two spinning lidars, the Ouster OS1-64 and the Ouster OS0-128. Table~\ref{table:sensor_details} shows the key specifications of these lidars, including the resolution of the images that they generate. It is worth noting that the vertical resolution of the images matches the number of channels in the lidar.

\begin{table}[t]
    \centering
    \caption{Lidar Specifications.} 
    \label{table:sensor_details}
    \resizebox{\columnwidth}{!}{
    \begin{tabular}{@{}lcccccccc@{}}  
        \toprule
        & Channels & FoV & Range & Freq. & Image resolution   \\
        \midrule   
        \textbf{Ouster OS1-64}      & 64 & 360°×45°     & 120\,m       & 10\,Hz  & 2048 x 128 \\ [+0.42em] 
        
        \textbf{Ouster OS0-128}     & 128  & 360°×90°      & 50\,m       & 10\,Hz  & 2048 x 64 \\ [+0.42em]  
        
        \textbf{RealSense L515}  &  N/A & 70°×55°           & 9\,m        & 30\,Hz  & 1920 x 1080  \\
        \bottomrule
    \end{tabular}
    }
\end{table}

% \begin{table}[H] 
%  \centering
%  \resizebox{\columnwidth}{!}{%
%     \begin{tabular}{@{}lcccccccc@{}}  
%     \toprule
%         & IMU & Type & Channels & FoV & Range & Freq. & Image resolution   \\
%     \midrule   
%         \textbf{Ouster OS1-64}      &  ICM-20948 & spinning & 64 & 360°×45°   & V:0.7°, H:0.18°    & 120\,m       & 10\,Hz  & 1,310,720 pts/s \\ [0.5ex] 
 
%         \textbf{Ouster OS0-128}      & ICM-20948 & spinning & 128  & 360°×90°    & V:0.7°, H:0.18°    & 50\,m       & 10\,Hz  & 2,621,440 pts/s \\ [0.5ex]  
        
%         \textbf{RealSense L515}  &  N/A & lidar camera & N/A & 70°×55°          & N/A       & 9\,m        & 30\,Hz  & -  \\
%      \bottomrule
%     \end{tabular}
% }   \vspace{0.2cm}
%     \caption{Lidar Specifications.} 
%      \label{table:sensor_details}
% \end{table}

Figure~\ref{fig:hardware_frame} depicts the data collection platform that can be mounted on different mobile platforms. The two lidars are installed on the sides, while an Intel RealSense L515 lidar camera captures RGB images.

\subsection{Data Acquisition}

We gathered data in various settings, including indoors and outdoors, day and night. For this initial assessment of the performance of DL models on images generated by the lidar sensors, we concentrate on a selection of object categories. These categories have been chosen based on the typical needs of autonomous systems as well as on objects that appear more often in the collected data. Outdoors, we analyze the detection of cars, bicycles and persons. Indoors, persons and chairs. Table~\ref{tab:data_instances} shows the number of object instances in the collected data. Samples of the data generated by the sensors are shown in Fig.~\ref{fig:data_samples}. In these examples, the resolution of a lidar-generated image is $2048 \times 128$ with $360^\circ$ field of view of a scene around while an RGB image of the L515 is $1920 \times 1080$.

% When we collect data, we concentrate on a few primary categories important in robotics and autonomous systems, such as people, automobiles, chairs, bikes, and similar objects. Figure~\ref{fig:data_samples} shows what the data we collected looks like. The lidar generated image is of $2048 \times 128$ with $360^\circ$ of the scene around while the image of L515 is the size of $1920 \times 1080$.

\begin{table}[t]
    \centering
    \caption{Instances of the different objects in the analyzed dataset}
    \label{tab:data_instances}
    \begin{tabular}{@{}l@{\hspace{2em}}c@{\hspace{1em}}c@{\hspace{2em}}c@{\hspace{1em}}c@{\hspace{1em}}c@{}}
        \toprule
         & \multicolumn{2}{c}{Indoors} & \multicolumn{3}{c}{Outdoors} \\
         & Person & Chair & Person & Car & Bike \\
        \midrule
        Instances & 43 & 42 & 103 & 37 & 14 \\
        \bottomrule
    \end{tabular}
\end{table}

\subsection{Data Preprocessing}

Regarding data preprocessing, we performed two main steps: denoising and interpolation. We considered different denoising and interpolation approaches and repeatedly ran object detection and segmentation on a set of test images. In our experiments, we applied a box filter to denoise the images and linear interpolation methods to properly resize the images.

\subsection{Object Detection Approaches}

Over the last decade, deep neural network models have achieved significant advances in computer vision, especially object detection. Object detection, which includes both object recognition and localization, is generally divided into two types: one-stage and two-stage detectors~\cite{zou2019object}. In this study, some of the most commonly used models from both frameworks are utilized for object detection.

\subsubsection{Two-Stage Object Detection}

A two-stage detector divided the detection process into a region proposal and classification phases. At the region proposal phase, several object candidates are proposed as regions of interest (RoI), classified and localized at the second phase. Object localization and detection are typically more accurate in models with a two-stage architecture than others. Two popular two-stage detectors were used in this study: FasterR-CNN~\cite{ren2015faster} and MaskR-CNN~\cite{he2017mask}. These models are implemented based on Pytorch, and ResNet-50 is used as the pre-trained backbone for object detection.

\subsubsection{One-Stage Object Detection}

In contrast to two-stage models, one-stage detectors utilize a single feed-forward fully-convolutional network for object feature extraction, bounding-box regression, and classification. In the one-stage approach, feature maps are detected and classified simultaneously. In addition to their excellent accuracy, the one-stage detector models are popular in real-time applications due to their high detection speed. One of the first widely adopted one-stage detectors in the deep learning field was YOLO, which was introduced in~\cite{redmon2016you}. Two variations of the YOLO model were applied in this study: YOLOx~\cite{ge2021yolox} and YOLOv5~\cite{jocher2020yolov5}. In the YOLOx toolset, there are different types of networks, including the YOLOx-s, YOLOx-m, YOLOx-l, and YOLOx-x models. We use the YOLOx-m model in this paper due to its high detection speed and performance.

\subsection{Image Segmentation Approaches}

Object segmentation is the process of assigning each pixel value of an image to a specific class and is generally divided into two types: semantic segmentation and instance segmentation. The semantic segmentation method considers objects that belong to the same class as a single group~\cite{garcia2018survey}, while the instance segmentation method combines semantic segmentation and object detection approaches and identifies multiple objects of a single class as distinct instances~\cite{hafiz2020survey}.

For semantic segmentation, HRNet + OCR + SegFix (High-Resolution Network) which placed 1st in the Cityscapes competition at ECCV 2020, is used~\cite{yuan2019segmentation}. Additionally, Pointrend~\cite{kirillov2020pointrend} and Mask R-CNN, both with ResNet-50 as their backbone, are employed, for instance segmentation.

% PointRend \\
% Mask-RCNN\\
% \input{sec/04_Methodology}
% \newpage

\begin{figure}
    \centering
    \begin{subfigure}{.95\columnwidth}
        \includegraphics[width=\columnwidth,height=0.2\textwidth]{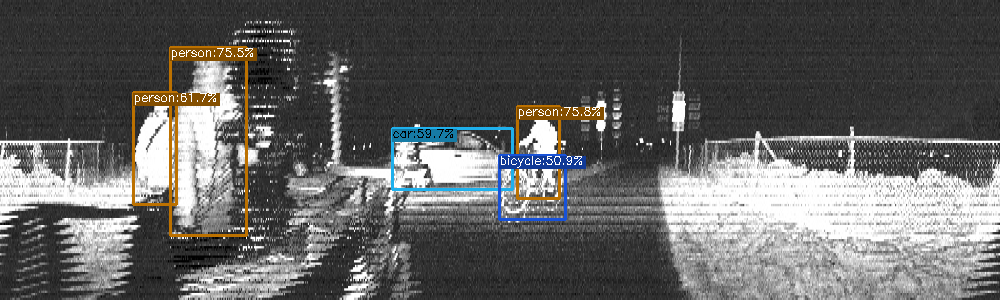}
        \caption{YOLOx detections in an outdoor scene}
        \label{fig:yolox_0}
    \end{subfigure}
    
    \vspace{1em}
    \begin{subfigure}{.95\columnwidth}
        \includegraphics[width=\columnwidth,height=0.2\textwidth]{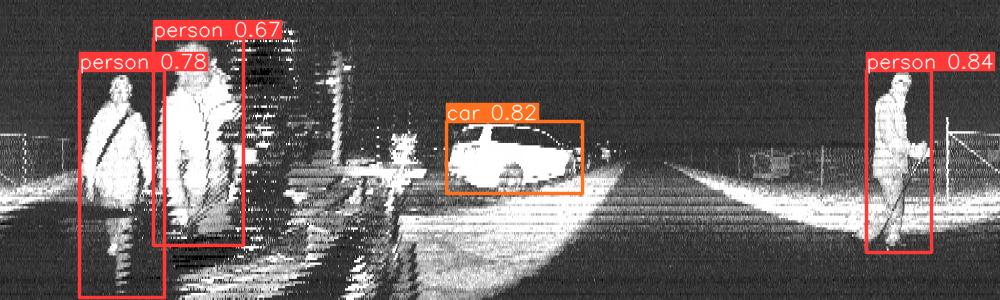}
        \caption{YOLOv5 detections in an outdoor scene} 
        \label{fig:yolov5_0}
    \end{subfigure}%
    
    \vspace{1em}
    \begin{subfigure}{.95\columnwidth}
        \includegraphics[width=\columnwidth,height=0.2\textwidth]{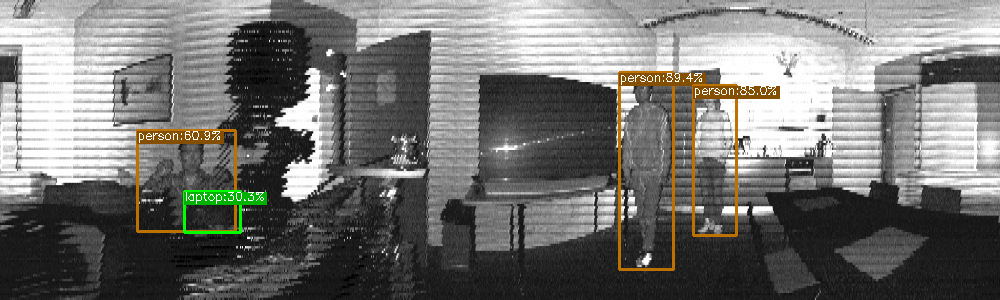}
        \caption{YOLOx detections in an indoor scene}
        \label{fig:yolox_1}
    \end{subfigure}%
    \caption{Detection examples in indoor and outdoor scenarios}
    \label{fig:detection_example}
\end{figure}

\section{Experimental Results}

Through this section we cover the results of applying the different object detection and instance segmentation models to the data gathered in the different environments.

\subsection{Detection Results}

The first part of the analysis delves into the performance of the different objectors.

% In the experimental results section, we evaluated the performance of multiple DL-based object detectors in general and more specific classes in particular. Additionally, we showed some object detection samples based on our lidar-generated $360^\circ$ image data to reference the performance of the DL algorithms.

% \subsubsection{Proportion of detected instances}

Table~\ref{tab:detection_accuracy} shows the proportion of objects successfully detected by Faster R-CNN, Mask R-CNN, YOLOv5, and YOLOx. Among them, YOLOx has a higher proportion of detected objects indoors and outdoors. It is worth noting that all four models were able to detect over 80\% of the persons indoors and over 80\% of the cars outdoors. In general, the performance of all the models is good enough to consider the adoption of this type of object detection in systems where the lidars are already present.

\begin{table}[t]
    \centering
    \resizebox{\columnwidth}{!}{
    \begin{tabular}{@{}l@{\hspace{1.5em}}l@{\hspace{1.5em}}c@{\hspace{1.5em}}c@{\hspace{1.5em}}c@{\hspace{1.5em}}c@{}}
        \toprule
        & & \textbf{Faster R-CNN} & \textbf{Mask R-CNN} & {\textbf{YOLOv5}} & {\textbf{YOLOx}} \\
        % & & \textbf{Faster-RCNN} & \textbf{Mask-RCNN} & \multirow{2}{*}{\textbf{YoloV5}} & \multirow{2}{*}{\textbf{YoloX}} \\
        % & & \textbf{RCNN} & \textbf{RCNN} & \textbf{} & \textbf{} \\ 
        \midrule
        {\multirow{2}{*}{\textbf{Indoors}}}   & Person    & 0.837 & 0.837 & 0.924 & \textbf{0.953} \\
                                             & Chair     & 0.357 & 0.333 & 0.398 & \textbf{0.515} \\[+0.6em]
        {\multirow{3}{*}{\textbf{Outdoors}}}  & Person    & 0.524 & 0.485 & 0.630 & \textbf{0.633} \\
                                           & Car       & 0.865 & 0.811 & \textbf{0.893} & 0.866 \\
                                           & Bike      & 0.357 & \textbf{0.643} & 0.143 & \textbf{0.571} \\ %[+1em]
        \bottomrule
    \end{tabular}
    }
    \vspace{0.2cm}
    \caption{Proportion of objects successfully detected by each of the models studied in this work. This metric does not include false negatives or false positives.}
    \label{tab:detection_accuracy}
\end{table}

For more specific metrics, in Table~\ref{tab:precision_recall} we show the precision and recall of the detectors. YOLOx has the most robust overall performance from the four different tested models.

\begin{table}[t]
    \centering
    \scriptsize
    \resizebox{\columnwidth}{!}{
    \begin{tabular}{@{}l@{\hspace{0.42em}}l@{\hspace{1.5em}}c@{\hspace{1em}}c@{\hspace{1.5em}}c@{\hspace{1em}}c@{\hspace{1.5em}}c@{\hspace{1em}}c@{\hspace{1.5em}}c@{\hspace{1em}}c@{}}
        % \footnotesize
        \toprule
        & & \multicolumn{2}{c}{\hspace{-1.23em}\textbf{Faster R-CNN}} & \multicolumn{2}{c}{\hspace{-1.23em}\textbf{Mask R-CNN}} & \multicolumn{2}{c}{\hspace{-1.23em}\textbf{YOLOv5}} & \multicolumn{2}{c}{\hspace{-1.23em}\textbf{YOLOx}}  \\
        & & Precision & Recall & Precision & Recall & Precision & Recall & Precision & Recall \\ 
        \midrule
        {\multirow{2}{*}{\textbf{In}}}   & Person      & 0.72 & 0.837 & 0.95 & 0.905 & 0.976 & 0.930 & \textbf{1.0} & \textbf{0.953} \\
                                             & Chair        & 1.0 & 0.115 & 0.57 & \textbf{0.826} & 1.0 & 0.115 & \textbf{1.0} & 0.315 \\[+0.6em]
        {\multirow{3}{*}{\textbf{Out}}}  & Person      & 0.912 & 0.505 & 0.957 & 0.464 & 0.872 & 0.854 & 0.969 & 0.653 \\
                                           & Car        & 0.943 & 0.688 & 0.712 & 0.627 & 0.919 & 0.829 & 0.825 & 0.618 \\
                                           & Bike            & 0.357 & 1.00 & 0.643 & 1.00 & 0.143 & 1.00 & 0.571 & 1.00 \\
        \bottomrule
    \end{tabular}
    }
    \vspace{0.2cm}
    \caption{Detection accuracy of multiple representative object detection networks in various scenarios}
    \label{tab:precision_recall}
\end{table}

Some other categories, including stop signs, handbags, and fire hydrants, are considered in our initial evaluation. However, they are not listed in Table~\ref{tab:detection_accuracy} and Table~\ref{tab:precision_recall} as we have focused on better analyzing a specific subset. In general, we have observed that both YOLOv5 and YOLOx can achieve comparable accuracy in these other classes as well.

\begin{figure*}
    \centering
    \begin{subfigure}{.48\textwidth}
        \includegraphics[width=\columnwidth,height=0.35\textwidth]{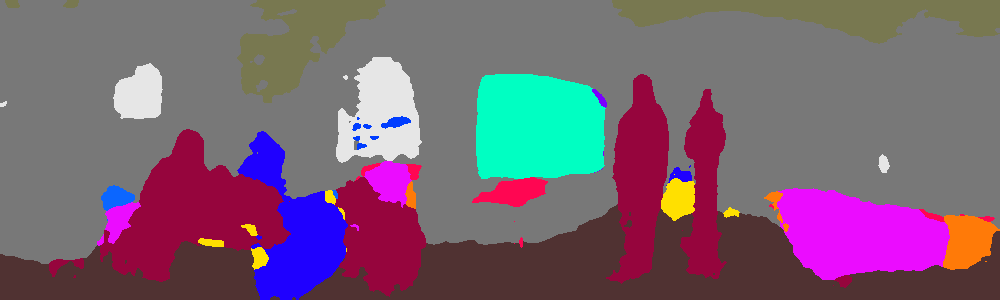}
        \caption{HRNet: Indoor Example}
        \label{fig:hrnet_indoors}
    \end{subfigure}%
    \hfill
    \begin{subfigure}{.48\textwidth}
        \includegraphics[width=\columnwidth,height=0.35\textwidth]{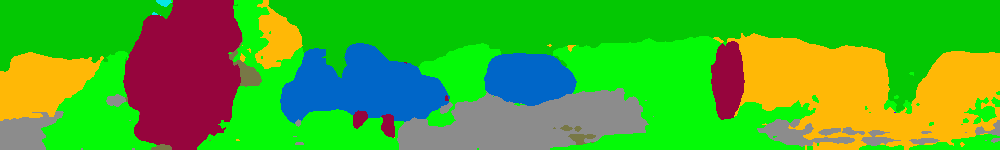}
        \caption{HRNet: Outdoor Example}
        \label{fig:hrnet_outdoors}
    \end{subfigure}%
    \caption{Indoor and Outdoor Semantic Segmentation Examples Based on HRNet}
    \label{fig:seg_examples}
\end{figure*}

% \subsubsection{Detection Examples on Lidar Generated Images}

In Fig.~\ref{fig:detection_example}, we show a sample of detection examples from YOLOv5 and YOLOx for both indoors and outdoors scenes.

% Detection: yolov5, yolox, faster-rcnn, mask-rcnn-resnet50-fpn, retinanet\_resnet50\_fpn

% Segmentation: maskrcnn, Lawin+, instance\_pointrend,  Cascade Eff-B7 NAS-FPN

% outdoors:
% 23,24,58,75,76,77,97,98,112,133,134,136,137,141, 143,148,150, 160, 217, 218, 229, 240, 307,
% 308, 311, 351, 364, 366, 380, 379, 415, 416, 421, 422, 435, 441
% indoors, ict-city 5th floor:
% 9,85, 94, 95, 276, 351, 352, 367, 371, 490, 

\subsection{Segmentation Results}

% \subsubsection{Semantic Segmentation Examples on Lidar Generated Images}

\begin{figure}
    \centering
    \begin{subfigure}{.95\columnwidth}
        \includegraphics[width=\textwidth,height=0.3\textwidth]{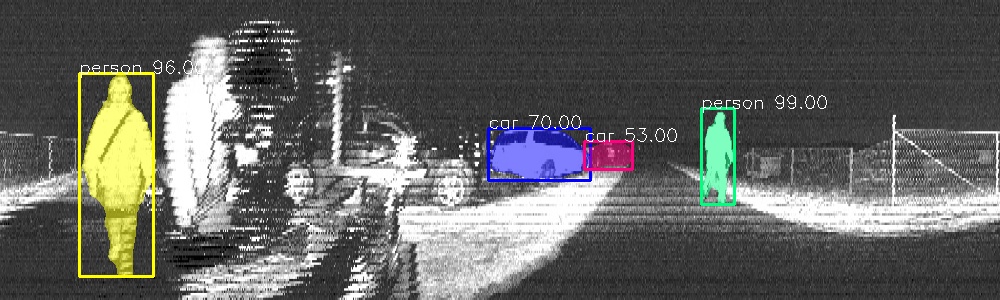}
        \caption{PointRend}
        \label{fig:Pointrend}
    \end{subfigure}%
    
    \vspace{1em}
    \begin{subfigure}{.995\columnwidth}
        \includegraphics[width=\textwidth,height=0.3\textwidth]{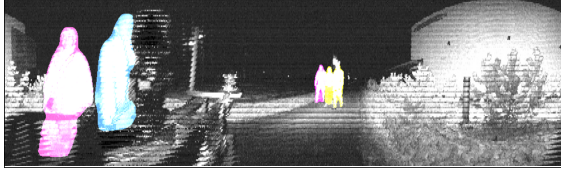}
        \caption{Mask R-CNN}
        \label{fig:Maskrcnn}
    \end{subfigure}%
    \caption{Indoor and Outdoor Instance Segmentation Examples Based on PointRend and Mask R-CNN}
    \label{fig:instance_examples}
\end{figure}

Regarding the performance of instance segmentation models, we show in Fig.~\ref{fig:seg_examples} shows examples of HRNet semantic segmentation in both indoor and outdoor scenes included. In Fig.~\ref{fig:instance_examples}, we also show the examples of instance segmentation results with PointRend and Mask R-CNN. In this case, the analysis is qualitative, and further results are available in the project's repository\footnote{https://github.com/TIERS/lidar-as-a-camera}. Nonetheless, our tests show good performance for the most typical object classes based on analysing a broad series of images.

% \subsubsection{Instance Segmentation on Lidar Generated Images}

\section{Conclusion and Future Work}

We have presented in this work an analysis of the performance of different object detection and semantic segmentation DL models on images generated by lidar sensors. We have collected data with two different lidars indoors and outdoors and in both daylight and night scenes. Our experiments show that state-of-the-art DL models can process this type of data with a promising performance by interpolating the low-resolution images to adequate resolutions. Object segmentation results are particularly optimistic, therefore paving the path for further usage of lidars beyond the current algorithms focused around odometry, localization, mapping and object detection from geometric methods.

In future work, we will explore a wider variety of preprocessing techniques and study the performance benefits of re-training some of the studied network architectures with data from the lidar camera sensors.

%%%%%%%%%%%%%%%%%%%%%%%%%%%%%%%%%%%%%%%%%%%%%%
%%                                          %%
%%            ACKNOWLEDGMENT                %%
%%                                          %%
%%%%%%%%%%%%%%%%%%%%%%%%%%%%%%%%%%%%%%%%%%%%%%

\section*{Acknowledgment}

This research work is supported by the R3Swarms project funded by the Secure Systems Research Center (SSRC), Technology Innovation Institute (TII).

%%%%%%%%%%%%%%%%%%%%%%%%%%%%%%%%%%%%%%%%%%%%%%
%%                                          %%
%%              BIBLIOGRAPHY                %%
%%                                          %%
%%%%%%%%%%%%%%%%%%%%%%%%%%%%%%%%%%%%%%%%%%%%%%
% \newpage
% \nocite{*}
\bibliographystyle{unsrt}
\bibliography{bibliography}

\end{document}